\documentclass[conference]{IEEEtran}
\IEEEoverridecommandlockouts
\PassOptionsToPackage{hyphens}{url}
\usepackage{graphicx}
\usepackage{caption}
\usepackage{tabularx}
\usepackage{listings}
\usepackage{hyperref}
\usepackage{bm}
\usepackage{booktabs}
\usepackage{xcolor}
\usepackage{amsmath}
\usepackage{amssymb}
\usepackage{cleveref}

\usepackage{xcolor}
\crefname{lstlisting}{listing}{listings}
\Crefname{lstlisting}{Listing}{Listings}

\newcommand{\teaserFigure}{%
  \centering
  \includegraphics[width=1\linewidth]{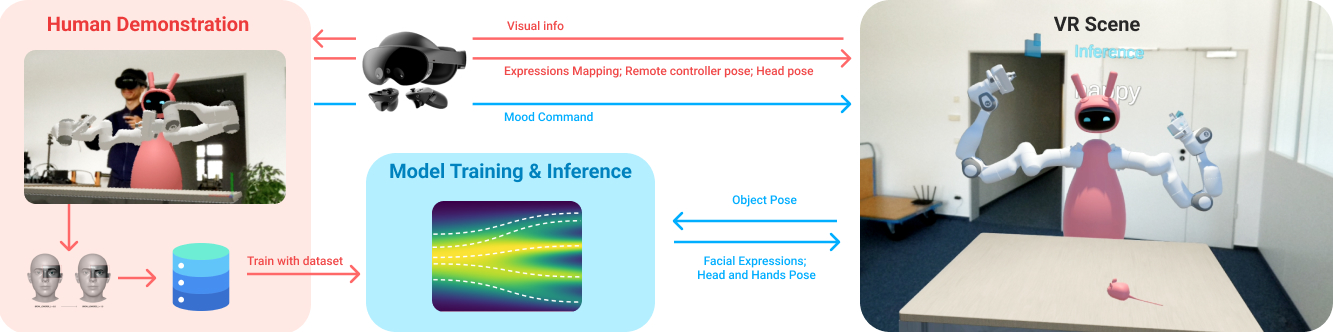}%
  \captionof{figure}{
Mixed-reality pipeline for learning robotic emotions.
(left) An expert wearing an HMD teleoperates a virtual robot; the system records facial blend-shapes together with head and hand/controller poses, forming an affect-rich demonstration dataset.
(centre) These demonstrations train a flow-matching generative model that maps a desired mood label plus live perceptual cues (the mouse pose) to continuous joint targets.
(right) At inference the trained model runs at 120 Hz, taking the operator’s high-level mood command and the pose of salient objects to drive the robot’s eyes, ears, neck and arms with recognisable emotions inside the MR scene.
Red (pink) arrows denote signals used only during data collection/training; blue arrows denote signals present at runtime.
}
  \label{fig:teaser}%
}

\makeatletter
\apptocmd{\@maketitle}{%
  \setcounter{figure}{0}% make sure the first visible caption is Fig. 1
  \teaserFigure\par      % show the teaser exactly once
}{}{}
\makeatother
\def\BibTeX{{\rm B\kern-.05em{\sc i\kern-.025em b}\kern-.08em
    T\kern-.1667em\lower.7ex\hbox{E}\kern-.125emX}}
\begin{document}

\title{Generation of Real-time Robotic Emotional Expressions Learning from Human Demonstration in Mixed Reality}

\newcommand{\asimo}{\includegraphics[scale=0.08]{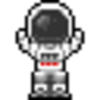}}

\author{Chao Wang\textsuperscript{\asimo}, 
        Michael Gienger\textsuperscript{\asimo},
        Fan Zhang\textsuperscript{\asimo}
%\thanks{*This work was not supported by any organization}% <-this % stops a space
\thanks{\asimo~\textit{Honda Research Institute Europe}, Germany
        {\tt\small \{firstname.lastname\}@honda-ri.de}}
        }
\maketitle
\begin{abstract}
Expressive behaviors in robots are critical for effectively conveying their emotional states during interactions with humans. In this work, we present a framework that autonomously generates realistic and diverse robotic emotional expressions based on expert human demonstrations captured in Mixed Reality (MR). Our system enables experts to teleoperate a virtual robot from a first-person perspective, capturing their facial expressions, head movements, and upper-body gestures, and mapping these behaviors onto corresponding robotic components including eyes, ears, neck, and arms. Leveraging a flow-matching-based generative process, our model learns to produce coherent and varied behaviors in real-time in response to moving objects, conditioned explicitly on given emotional states. 
A preliminary test validated the effectiveness of our approach for generating autonomous expressions.
Supplementary material can be found at \url{https://wallacewangchao.github.io/fm-expressions/}

\end{abstract}

\section{Introduction}
Expressive behaviour is a cornerstone of engaging human-robot interaction (HRI): people interpret a robot’s gaze, posture and motion as cues to its internal state, and emotionally expressive robots are trusted more and even forgiven for mistakes more readily than impassive ones \cite{breazeal2009role, bretan2015emotionally, liu2023robots}. Designing such behaviour, however, is hard. Hand-crafted rule sets do not scale to the diversity of human affect, and supervised pipelines demand large, carefully annotated datasets that are expensive to gather and struggle to generalise in real time ~\cite{mahadevan2024generative, stock2022survey}.
Recent work has begun to replace rigid scripts with learned generative models that synthesise motion directly from high-level intent, yielding richer expressions but still relying on offline generation or heavy post-processing 
~\cite{zhang2025exface,elegnt-expressive-functional-movement}. In parallel, Mixed-Reality (MR) teleoperation has emerged as an effective way to capture nuanced human demonstrations—including subtle facial cues and upper-body gestures—without constraining the expert’s viewpoint or embodiment 
~\cite{black2024mixed}. At the modelling level, flow-matching generative processes promise fast, stable synthesis and naturally accommodate continuous control, making them attractive for real-time robots 
~\cite{lipman2022flow,zhang2024affordance}.
\begin{figure*}[!t]
    \centering
    \includegraphics[width=1\linewidth]{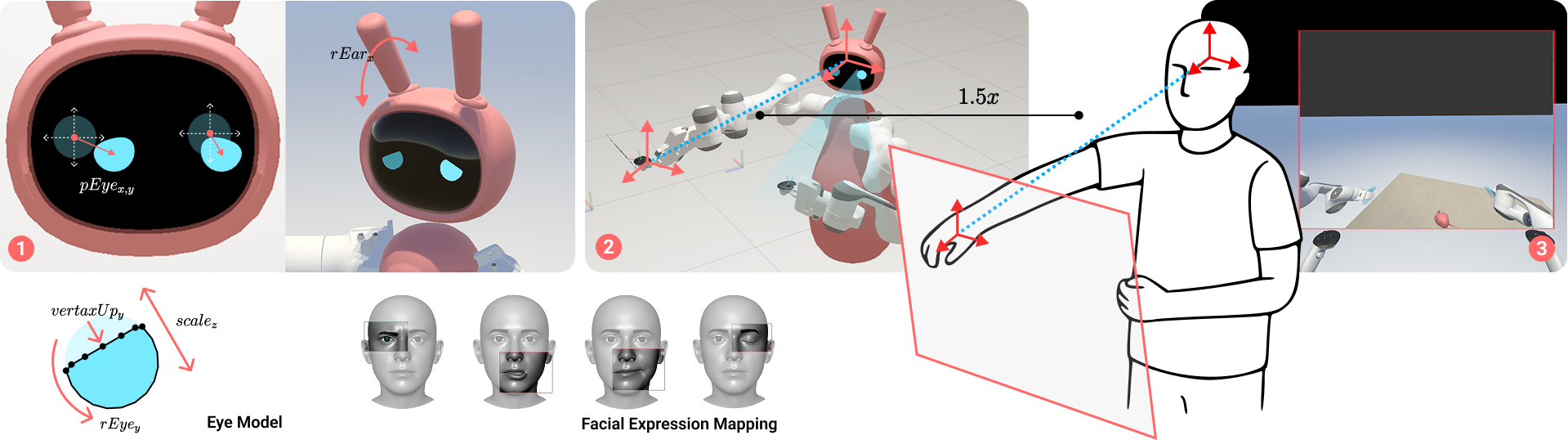} 
    \caption{The XR platform: 1. 7 facial-expression values detected by the XR-headset map the robot's ears angle and shape of the eyes, the gaze direction also maps the position of the eyes on the plane of robot face screen. Some value of the facial expression also maps the movement of the robot's ear. 2. Human's head position and orientation maps the robot's end effector, relative to the operator's head pose as the origin. The positional value is sceled by 1.5 for enhancing operator's reachablitiy. 3. There is an virtual screen floating in front of the operator, which allows the operator to observe the environment from the first person perpective.}
    \label{fig:xr-platform}
\end{figure*}

We unite these strands and present a framework that autonomously produces diverse and coherent emotional expressions on a physical robot in real time. Experts teleoperate a virtual robot in MR from a first-person perspective; their facial expressions, head motions and body gestures are mapped onto robotic eyes, ears, neck and arms, providing rich demonstrations. A flow-matching generator then learns to conditioning a desired emotional label and live perceptual cues (e.g., a moving object) into continuous joint poses, yielding behaviour that observers recognise as naturally “fear”, “angry”, “curious”, "sad",  "bored" and or “happy” at 10 Hz. 
% In a user study, participants identified our autonomous expressions as reliably as those produced by direct expert teleoperation.
The contributions of this study are listed as follows:
\begin{enumerate}
    \item MR demonstration pipeline: a device-agnostic capture method that records fine-grained affective motion from a first-person operator.
    \item Flow-matching emotional generator: the first application of flow matching to real-time robotic expression, conditioning on explicit emotion labels.
\end{enumerate}

\section{Our Approach}
Our framework is composed of two essential components: a platform for gathering training data and a real-time inference system for emotional expressions.

\subsection{Mixed--Reality Data-Collection Platform}

Our data are gathered in an XR application built in \textit{Unity} using the Meta Quest Developer SDK~(v74) and deployed on a Quest Pro headset.  
When the app launches, a mixed-reality scene appears in which a mobile bimanual robot (base, torso, Kinova arms, head, LED eyes, and actuated ears) is placed on the real floor in front of the operator (Fig.~\ref{fig:teaser}, right).  
A table and a small \emph{mouse} prop that scurries along random trajectories are spawned between the robot and the user.  
The moving mouse serves as a salient target that elicits richer, direction-specific expressions than a static cue.  
Because passthrough video keeps the real surroundings visible, the operator retains situational awareness while demonstrating (Fig.~\ref{fig:teaser} right).

\subsubsection{End effector mapping}
The Quest headset provides 6-DoF poses for the user’s head and for each hand-held controller.  
The robot base is fixed; hence we map only the \emph{orientation} of the operator’s head to the robot head, discarding translation.  
For the arms we adopt a relative mapping: the position of each controller expressed in the operator’s head frame is scaled by~1.5 and retargeted to the corresponding robot end-effector, while orientations are matched directly.  
The scale factor compensates for the robot’s greater reach, allowing the user to command poses that lie outside their own workspace (step 2 in Fig ~\ref{fig:xr-platform}).

\subsubsection{Facial-Expression Mapping}
Quest Pro face tracking outputs 70 per-muscle blend-shape values in~$[0,1]$.\footnote{\url{https://developers.meta.com/horizon/documentation/unity/move-face-tracking}}  
We select seven that most affect perceived emotion:  
\emph{eye closedness} $C_{\text{eye}}$, \emph{lip-corner dimple} $D_{\text{lip}}$, \emph{brow lower} $H_{\text{brow}}$ (all left and right), and \emph{chin raise} $H_{\text{chin}}$.  
Gaze direction $(\theta_x,\theta_y)$ is also available.
The robot eye consists of two cylinders whose shapes, scales, and positions are controlled, and each ear has one rotational DoF.  Let
$\text{vertaxUp}_y$, $\text{vertaxLow}_y$, $r_{\text{Eye}z}$, $s_{\text{Eye}y}$, $r_{\text{Ear}x}$, and $(p_{\text{Eye}x},p_{\text{Eye}y})$ denote those DoFs (step 1 in ~\ref{fig:xr-platform}).  
We employ the mapping:
\begin{align*}
\text{vertaxLow}_y &= \min\!\bigl(D_{\text{lip}},\;\text{vertaxLow}_y\bigr),\\
\text{vertaxUp}_y  &= \max\!\bigl(-\tfrac{H_{\text{chin}}+H_{\text{brow}}}{2},\;\text{vertaxUp}_y\bigr),\\
r_{\text{Eye}y}     &= (H_{\text{chin}}+H_{\text{brow}})\,\pi/6,\\
r_{\text{Ear}x}     &= \pi/2\,(-H_{\text{chin}}+H_{\text{brow}}),\\
s_{\text{Eye}z}     &= 1-0.9\,C_{\text{eye}},\\
(p_{\text{Eye}x},p_{\text{Eye}y}) &= \operatorname{clamp}\!\bigl(-(\theta_x,\theta_y)/\theta_{\max},\,-1,1\bigr).
\end{align*}

\subsubsection{First-Person View} 
A virtual monitor rigidly attached to the headset shows live footage from a camera mounted on the robot’s head, giving the operator a first-person preview of how each demonstrated motion will appear in situ.  
Pressing either controller’s trigger sends the current observation–action pair to a back-end server for logging, and the screen border flashes red as confirmation (step 3 in ~\ref{fig:xr-platform}).

\subsubsection{WebSocket Back-End}
A lightweight WebSocket server streams observations to disk at ${10}{Hz}$ during demonstration and, later, relays live observations to the trained model for closed-loop inference.

\subsection{Hand Mapping}

\subsection{Real Robot Implementation}
\begin{figure}
    \centering
    \includegraphics[width=\linewidth]{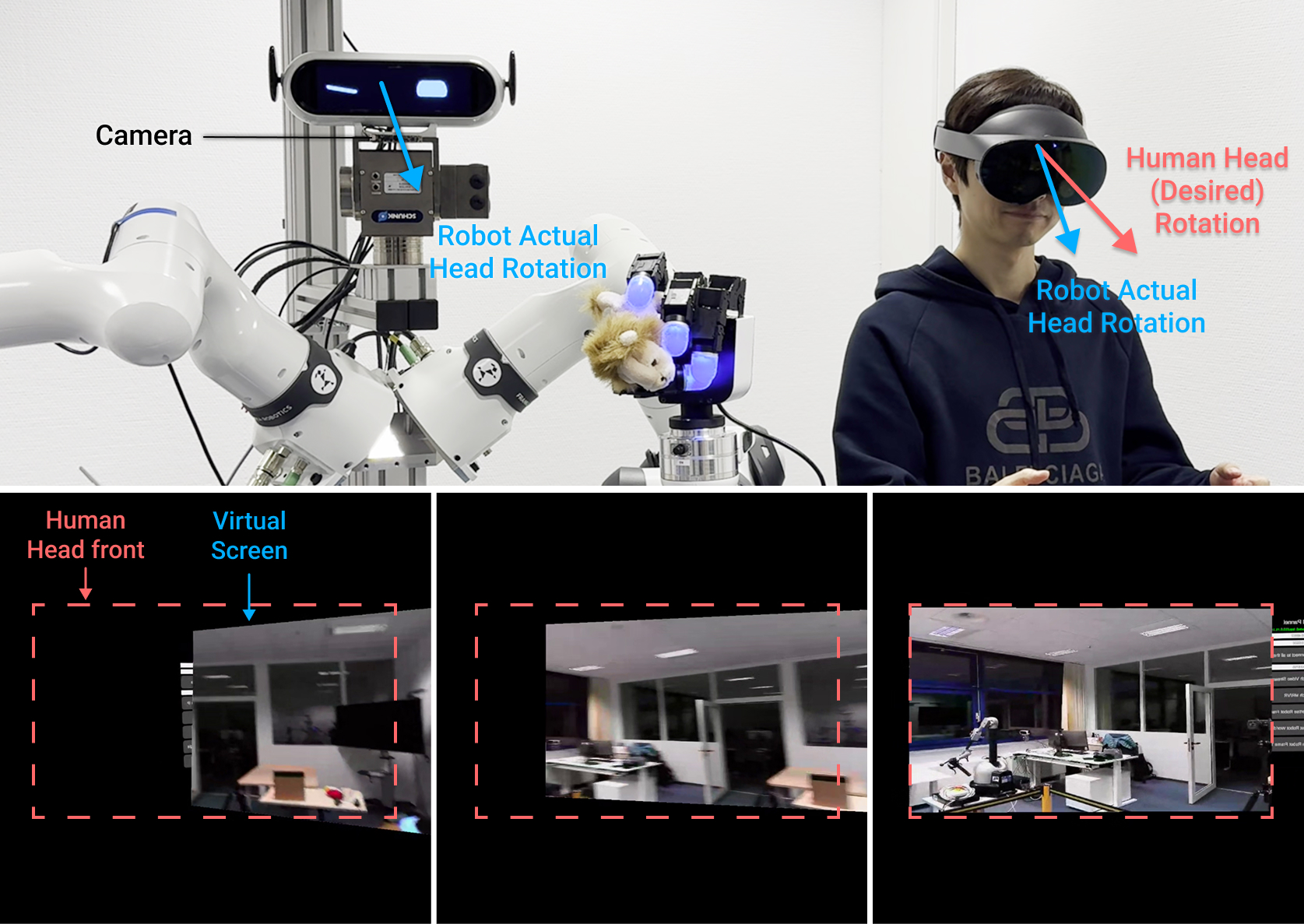} 
    \caption{
    \textbf{Top}: The MR headset subscribes to the actual rotation of the robot head and computes the rotation difference with respect to the human head.
    \textbf{Bottom}: The teleoperation view locally compensates the virtual screen orientation according to this difference, aligning the displayed video stream with the robot head pose and reducing visual mismatch to mitigate motion sickness.
    }
    \label{fig:real-screen}
\end{figure}

Our approach can be directly integrated into a real robotic system for data collection and teleoperation.  
The operator’s head pose and facial expression data are transmitted to the robot using the same mapping strategy described in the previous sections.

The visual feedback is captured by a camera mounted on the robot’s neck and streamed to the operator through a WebRTC-based communication pipeline.
However, when deploying the system on a real robot, a critical challenge arises from the discrepancy between human head motion and robot neck dynamics.
Specifically, the robot neck typically responds more slowly than human head rotations.

If the virtual screen rigidly follows the human head pose while the robot head lags behind, the displayed visual content no longer corresponds to the robot’s actual viewing direction.
This inconsistency can lead to a strong visual–vestibular mismatch and significantly increase the risk of motion sickness for the operator.

To address this issue, the MR application continuously subscribes to the robot’s actual head rotation and computes its difference relative to the human head orientation.
Instead of fixing the virtual screen directly in front of the operator or strictly following the human head motion, the screen orientation is adjusted based on this rotation difference.
As a result, the virtual screen rotates more slowly than the human head, remaining aligned with the robot’s real viewing direction.

This adaptive compensation strategy not only reduces visual inconsistency and motion sickness, but also allows the operator to better perceive and anticipate the robot’s physical head movements during teleoperation.

\subsection{Flow-Matching Policy}
We develop the robot's behavioral cloning strategy as a generative mechanism using flow matching, which creates a flow vector to incrementally shift a source probability distribution into a target distribution. Unlike Denoising Diffusion Probabilistic Models (DDPM), which employ a stochastic differential equation by introducing noise, flow matching utilizes an ordinary differential equation to deterministically shape the data distribution.
\begin{figure}
    \centering
    \includegraphics[width=\linewidth]{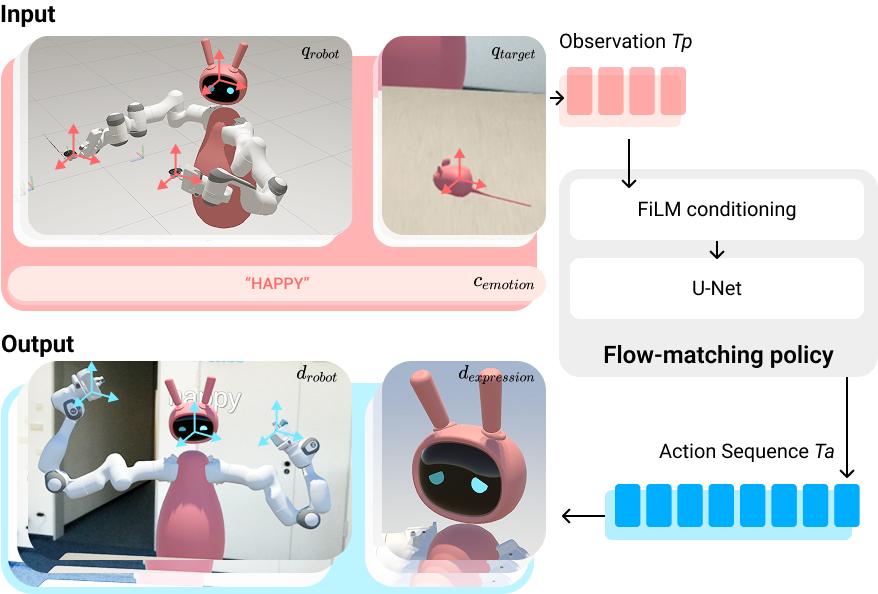}
    \caption{Overview of flow matching for expression generation. A history window of robot and target poses plus an emotion label (pink) is fed through FiLM-conditioned U-Net to predict the blue action sequence executed on the robot.}
    \label{fig:method-architechture}
\end{figure}
\subsubsection{Flow Matching Model}
Given a conditional probability density path $p_t(\bm x | \bm z)$ and a corresponding conditional vector field $\bm u_t(\bm x | \bm z)$, the objective loss of flow matching could be described as:
\begin{equation}
\mathcal{L}_{\text{FM}}(\bm \theta) = \mathbb{E}_{t, q(\bm z), p_t(\bm x | \bm z)} \left\| \mathbf{v}_t(\bm x, \bm \theta) - \mathbf{u}_t(\bm x | \bm z) \right\|^2
\label{eq:loss}
\end{equation}
where $\bm x \sim p_t(\bm x | \bm z)$, $t \sim \mathcal{U}[0, 1]$. Flow matching aims to regress $\mathbf{u}_t(\bm x | \bm z)$ with a time-dependent vector field of flow $\mathbf{v}_t(\bm x, \bm \theta)$ parameterized as a neural network with weights $\bm \theta$. $\mathbf{u}_t(\bm x | \bm z)$ can be further simplified as:
\begin{equation}
\mathbf{u}_t(\bm x | \bm z) = \bm x_1 - \bm x_0 \qquad \bm x_0 \sim p_0, \bm x_1 \sim p_1
\nonumber
\end{equation}
$p_0$ represents a simple base density at time $t=0$, $p_1$ denotes the target complicated distribution at time $t=1$, $\bm x_0$ and $\bm x_1$ are the corresponding samplings. $\mathbf{v}_t(\bm x, \bm \theta)$ is described as: 
\begin{equation}
\mathbf{v}_t(\bm x, \bm \theta) = v_{\bm \theta}(\bm x_t, t)
\label{eq:flow}
\end{equation}
where we define $\bm x_t$ as the linear interpolation between $ \bm x_0$ and $\bm x_1$ with respect to time $\bm x_t = t\bm x_1+(1-t)\bm x_0$, following the linear conditional flow theory~\cite{peyre2019computational}. $v_{\bm \theta}$ is a network of the flow model. Thus Equation~(\ref{eq:loss}) could be reformatted as 
\begin{equation}
\mathcal{L}_{\text{FM}}(\bm \theta) = \mathbb{E}_{t, \sim p_0, \sim p_1} \left\| v_{\bm \theta}(\bm x_t, t) -(\bm x_1 - \bm x_0)) \right\|^2
\label{eq:newloss}
\end{equation}
This represents the progression of the scalar flow that transforms data from source to target between time 0 and 1.

\textbf{Observation Conditioning and Generated Actions:} We modify Equation~(\ref{eq:flow}) to allow the model to predict actions conditioned on observations:
\begin{equation}
\mathbf{v}_t(\bm x | \bm{o}) = v_{\bm \theta}(\bm x_t, t| \bm{o})
\nonumber
\end{equation}
For the input to the flow matching model, we first use one-hot for emotional classification embeddings $c_{emotion}$, then concatenate with the 6D robot head and end hands' pose $q_{robot}$, and the target object pose $q_{target}$ that the robot is focusing on. The flow model $f_{\bm \theta}$ is represented with U-Net~\cite{ronneberger2015u}. The flow model predicts vectors $\mathbf{v}_t$ conditioned on the observation $\bm{o}$ with Feature-wise Linear Modulation (FiLM)~\cite{perez2018film}.

The output $\bm x$ here denotes robot actions including head and end hands' pose $d_{robot}$, and robot facial expression $d_{expression}$. $\bm x_1$ in Equation~(\ref{eq:newloss}) represents the demonstration robot actions. $\bm x_0$ is the random generated actions following a multivariate normal distribution $\bm x_0 \sim \mathcal{N}(0, I)$. 

\textbf{Closed-loop action prediction:} We execute the action prediction obtained by our flow matching model for a fixed duration before replanning. At each step, the policy takes the observation data $\bm o$ as input and predicts $Tp$ steps of actions, of which $Ta$ steps of actions are executed on the robot without re-planning. $Tp$ is the action prediction horizon and $Ta$ is the action execution horizon. 

\textbf{Inference:} For the inference procedure, random waypoints are sampled from the source distribution and then flowed into the target trajectory by estimating the flow from $t = 0$ to $t = 1$ over steps. We could use multiple steps $1/\Delta t$ for inference:
\begin{equation}
\bm x_{t+ \Delta t} = x_{t}+\Delta t f(\bm x_t, t| \bm{o}),  \qquad \text{for} \  t \in [0, 1]
\label{eq:inference}
\end{equation}

\begin{figure*}
    \centering
    \includegraphics[width=\linewidth]{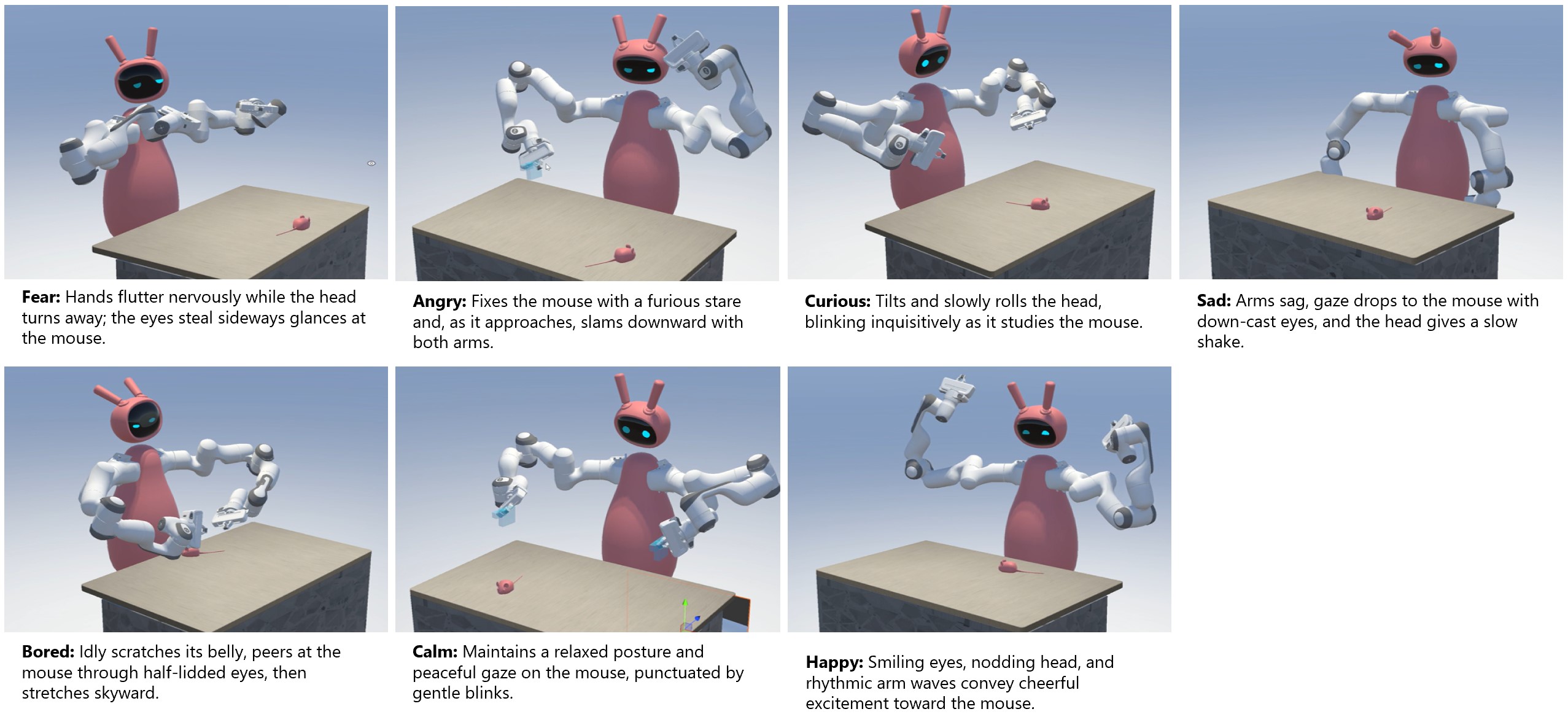} 
    \caption{Generated emotional expressions directed toward the moving target object.}
    \label{fig:emotion-pie}
\end{figure*}
\subsubsection{Emotional-Expression Dataset}
\label{sec:dataset}
We collected demonstrations for seven emotions
(\textit{happy, sad, angry, fear, bored, curious, calm})
towards the moving mouse target (Fig.~\ref{fig:emotion-pie}).
Each clip contains $\sim\!10\,000$ frames sampled at
${10}{Hz}$, logging robot poses, end-effector trajectories, gaze direction, and eye/ear motions.
The resulting dataset couples full-body motion with fine-grained facial animation under varied spatial context, providing rich supervision for the flow model.

\section{Preliminary Results and Future Directions}
\label{sec:results}

\textbf{Training protocol.}
The flow model was trained for ${3000}$ epochs with a batch size of~256 and a learning rate of~$1\!\times\!10^{-4}$.  
We explored four history-window lengths ($1$, $2$, $4$, and $16$ frames) in combination with two prediction horizons ($16$ and $32$ frames).

\textbf{Expert appraisal.}
A panel of HRI researchers informally inspected roll-outs and provided qualitative feedback:

\begin{enumerate}
\item \emph{Temporal context.}  
A \(16\)-frame history performed noticeably worse than \(2\)–\(4\) frames, suggesting that our FiLM-conditioned U-Net does not fully exploit long temporal correlations.  
Replacing FiLM with a transformer-based temporal encoder may improve sequence understanding at the cost of heavier training.

\item \emph{Prediction horizon.}  
Longer horizons (\(32\) frames) produced more complete, fluid gestures, whereas short horizons introduced occasional “jumps’’ when the policy re-planned.  
This points to a weak internal notion of phase; additional data or an explicit timing signal could reduce discontinuities.

\item \emph{Emotion coverage.}  
Six of the seven emotions transferred convincingly; the \textit{curious} behaviour lacked the distinctive “poke’’ motion present in the demonstrations.  
We attribute this to data sparsity and will extend the dataset with targeted examples.
\end{enumerate}

\textbf{Next steps.}
We will (i) integrate a transformer backbone for richer temporal reasoning, (ii) expand the training corpus to balance under-represented actions, and (iii) conduct a controlled user study to quantify recognisability, naturalness, and preference compared with teleoperation baselines.
\bibliographystyle{IEEEtran_etal}
\bibliography{references.bib}

@article{breazeal2009role,
  title={Role of expressive behaviour for robots that learn from people},
  author={Breazeal, Cynthia},
  journal={Philosophical Transactions of the Royal Society B: Biological Sciences},
  volume={364},
  number={1535},
  pages={3527--3538},
  year={2009},
  publisher={The Royal Society}
}

@article{bretan2015emotionally,
  title={Emotionally expressive dynamic physical behaviors in robots},
  author={Bretan, Mason and Hoffman, Guy and Weinberg, Gil},
  journal={International Journal of Human-Computer Studies},
  volume={78},
  pages={1--16},
  year={2015},
  publisher={Elsevier}
}

@article{liu2023robots,
  title={Robots’“woohoo” and “argh” can enhance users’ emotional and social perceptions: An exploratory study on non-lexical vocalizations and non-linguistic sounds},
  author={Liu, Xiaozhen and Dong, Jiayuan and Jeon, Myounghoon},
  journal={ACM Transactions on Human-Robot Interaction},
  volume={12},
  number={4},
  pages={1--20},
  year={2023},
  publisher={ACM New York, NY}
}

@inproceedings{mahadevan2024generative,
  title={Generative expressive robot behaviors using large language models},
  author={Mahadevan, Karthik and Chien, Jonathan and Brown, Noah and Xu, Zhuo and Parada, Carolina and Xia, Fei and Zeng, Andy and Takayama, Leila and Sadigh, Dorsa},
  booktitle={Proceedings of the 2024 ACM/IEEE International Conference on Human-Robot Interaction},
  pages={482--491},
  year={2024}
}

@article{stock2022survey,
  title={Survey of emotions in human--robot interactions: Perspectives from robotic psychology on 20 years of research},
  author={Stock-Homburg, Ruth},
  journal={International Journal of Social Robotics},
  volume={14},
  number={2},
  pages={389--411},
  year={2022},
  publisher={Springer}
}

@article{zhang2025exface,
  title={ExFace: Expressive Facial Control for Humanoid Robots with Diffusion Transformers and Bootstrap Training},
  author={Zhang, Dong and Peng, Jingwei and Jiao, Yuyang and Gu, Jiayuan and Yu, Jingyi and Chen, Jiahao},
  journal={arXiv preprint arXiv:2504.14477},
  year={2025}
}

@misc{elegnt-expressive-functional-movement,
title = {ELEGNT: Expressive and Functional Movement Design for Non-Anthropomorphic Robot},
author = {Yuhan Hu and Peide Huang and Mouli Sivapurapu and Jian Zhang},
year = {2025},
URL = {https://arxiv.org/abs/2501.12493}
}

@article{black2024mixed,
  title={Mixed reality human teleoperation with device-agnostic remote ultrasound: Communication and user interaction},
  author={Black, David and Nogami, Mika and Salcudean, Septimiu},
  journal={Computers \& Graphics},
  volume={118},
  pages={184--193},
  year={2024},
  publisher={Elsevier}
}

@article{lipman2022flow,
  title={Flow matching for generative modeling},
  author={Lipman, Yaron and Chen, Ricky TQ and Ben-Hamu, Heli and Nickel, Maximilian and Le, Matt},
  journal={arXiv preprint arXiv:2210.02747},
  year={2022}
}

@article{zhang2024affordance,
  title={Affordance-based Robot Manipulation with Flow Matching},
  author={Zhang, Fan and Gienger, Michael},
  journal={arXiv preprint arXiv:2409.01083},
  year={2024}
}

@inproceedings{ronneberger2015u,
  title={U-net: Convolutional networks for biomedical image segmentation},
  author={Ronneberger, Olaf and Fischer, Philipp and Brox, Thomas},
  booktitle={Medical image computing and computer-assisted intervention--MICCAI 2015: 18th international conference, Munich, Germany, October 5-9, 2015, proceedings, part III 18},
  pages={234--241},
  year={2015},
  organization={Springer}
}

@article{peyre2019computational,
  title={Computational optimal transport: With applications to data science},
  author={Peyr{\'e}, Gabriel and Cuturi, Marco and others},
  journal={Foundations and Trends{\textregistered} in Machine Learning},
  volume={11},
  number={5-6},
  pages={355--607},
  year={2019},
  publisher={Now Publishers, Inc.}
}

@inproceedings{perez2018film,
  title={Film: Visual reasoning with a general conditioning layer},
  author={Perez, Ethan and Strub, Florian and De Vries, Harm and Dumoulin, Vincent and Courville, Aaron},
  booktitle={Proceedings of the AAAI conference on artificial intelligence},
  volume={32},
  year={2018}
}
\end{document}